\documentclass[runningheads]{llncs}
\usepackage{graphicx}
\usepackage{hyperref}
\usepackage[english]{babel}

\title{NeuralSympCheck: A Symptom Checking and Disease Diagnostic Neural Model with \\ Logic Regularization}
\titlerunning{A Symptom Checking and Disease Diagnostic Neural Model}

\author{Aleksandr Nesterov\inst{1}\orcidID{0000-0003-1126-8099} \and
Bulat Ibragimov\inst{2}\orcidID{0000-0001-8540-0684} \and
Dmitriy	Umerenkov\inst{1}\orcidID{0000-0003-0413-7170} \and
Artem Shelmanov\inst{2,3}\orcidID{0000-0002-2151-6212} \and
Galina Zubkova\inst{1}\orcidID{0000-0001-9555-1689} \and
Vladimir Kokh\inst{1}\orcidID{0000-0002-9257-0259}}

\authorrunning{A. Nesterov et al.}

\institute{\textsuperscript{1}Sber AI Lab,
\textsuperscript{2}AIRI,
\textsuperscript{3}Skoltech 
\\
Moscow, Russia \\
\email{AINesterov@sberbank.ru}}

\begin{document}

\maketitle

\section{Abstract}

The symptom checking systems inquire users for their symptoms and perform a rapid and affordable medical assessment of their condition. The basic symptom checking systems based on Bayesian methods, decision trees, or information gain methods are easy to train and do not require significant computational resources. However, their drawbacks are low relevance of proposed symptoms and insufficient quality of diagnostics. The best results on these tasks are achieved by reinforcement learning models. Their weaknesses are the difficulty of developing and training such systems and limited applicability to cases with large and sparse decision spaces. We propose a new approach based on the supervised learning of neural models with logic regularization that combines the advantages of the different methods. Our experiments on real and synthetic data show that the proposed approach outperforms the best existing methods in the accuracy of diagnosis when the number of diagnoses and symptoms is large. The models and the code are freely available online\footnote{\url{https://github.com/SympCheck/NeuralSymptomChecker}}.

\textbf{Keywords:} Neural Networks, Symptom Checker, Diagnostic Model

\section{Introduction}

Health systems need to balance three critical qualities: accessibility, quality, and cost. These three qualities unfortunately often compete over a limited pool of resources, and improving one of these qualities leads to losses in others. This is known as the ``iron triangle'' of healthcare. Mobile networks, big data, and artificial intelligence are promising directions for improving quality and accessibility while decreasing costs. In \cite{semigran2015evaluation}, authors show that in 2012 35\% of adult US citizens at least once used the internet for self-diagnosis. Self-diagnosis commonly starts with queries to search engines. While highly accessible and free, the quality of the results may be unsatisfactory, and results may be irrelevant, inaccurate, or even harmful.

To increase the quality of self-diagnosis, several symptom-checker systems have been proposed \cite{semigran2015evaluation,tang2016inquire}. Such systems present users with several additional questions about existing or potential symptoms and use this information to suggest possible diagnoses and recommend visiting a specialist physician. The disease diagnosing process can be modeled as a sequence of questions and answers: a physician asks a patient questions about his/her symptoms and uses the answers to identify the disease. While asking the questions, the physician pursues two goals. Firstly, the answer to each question must be the most informative in the current context. Secondly, after a series of questions and answers, a correct diagnosis should be identified.

This work presents a symptom checker based on a logic regularisation framework \cite{asai2020logic}, which outperforms the state-of-the-art results achieved with methods based on reinforcement learning (RL). Unlike the RL systems, the proposed symptom checker is simple both in implementation and training. We split the system into symptom recommendation and diagnosis prediction submodels and implement the novel logic regularisation framework that allows us to train the submodels simultaneously with the standard backpropagation and to treat the symptom suggestion as a multi-label classification task. The latter, in turn, allows to deal with the problem of big and sparse symptom space by using the Asymmetric loss \cite{ridnik2021asymmetric}. In contrast to RL-based systems, the diagnoses predicted with our system also do not depend on the order of presented symptoms. The contributions of the paper can be summarized as follows:
\begin{itemize}
 \item We present a symptom checker that outperforms state-of-the-art systems based on reinforcement learning or knowledge graphs in the task of diagnosis prediction both on real world and synthetic datasets.
 \item Instead of the RL framework, we apply logic regularisation to train the symptom checker, showing that simpler models can achieve state-of-the-art results.
 \item Unlike the predictions of the RL-based diagnosis systems, the predictions of our system do not depend on the order of the revealed symptoms.
 \item Our system is easier to implement, train and requires less computational resources than state-of-the-art RL-based systems.
 \item Reframing the symptom recommendation problem as a multi-label classification task allows dealing with the big and sparse symptom space using the Asymmetric loss \cite{ridnik2021asymmetric}. 
\end{itemize}

\section{Related Work}

Early works concerning automated symptom clarification and diagnostics were based on the naive Bayes classifier, decision trees, and other information-gain methods \cite{kohavi1996scaling,kononenko2001machine}. Due to simplicity and various drawbacks, such systems do not achieve high diagnostics quality. There are also attempts to use rule-based expert systems \cite{hayashi1991neural}. The performance of such systems depends on the quality of the rules and medical knowledge bases. Therefore, scaling and modifying them is very difficult.

Several recent works \cite{tang2016inquire,wei2018task,kao2018context,peng2018refuel,janisch2020classification} demonstrate effectiveness of RL-based methods for these tasks. In the RL framework, the symptom clarification and diagnosis prediction tasks are framed as a Markov decision process \cite{tang2016inquire,wei2018task}. This 
%, unfortunately, 
leads to the unwanted quality of RL-based systems that the final diagnosis depends on the order in which the symptoms are revealed. Despite impressive results, the RL-based approach is plagued with several difficulties. Firstly, the possible symptoms and diagnoses are numbered from hundreds to a tenth of thousands, which leads to a huge decision space. Secondly, the number of symptoms present in each case is tiny compared to all possible symptoms, leading to a sparse decision space. To overcome this difficulties, in \cite{tang2016inquire,kao2018context}, the authors propose ensembling that helps to reduce the decision space for each individual ensemble component and improve the qualitative performance. Peng et al. \cite{peng2018refuel} addresses the sparsity of decision space by proposing special reward estimation and regularisation techniques. To increase the diagnostic performance, other models use context (age, sex, location) \cite{kao2018context} or information from knowledge graphs \cite{zhao2020weighted}.

In \cite{he2020fit,lin2020learning}, the authors note that the decision to stop the dialog made by an RL-agent can be sub-optimal because the agent is penalized for long conversations. To solve the problem, they use uncertainty estimation \cite{mcallister2019robustness} of the diagnosis as a stopping criteria. The quality of diagnostics improves, because the agent makes more steps, and the diagnostics model receives more information. 

Our logic regularisation framework is similar to RL-based methods as it both models the dialog between the physician and the patient and achieves high-quality results. At the same time, our system is easier to implement and train, and its predictions are independent of the order of revealed symptoms. 

\begin{figure}[t]

\center{

\scalebox{1.2}{
\includegraphics[scale=0.5]{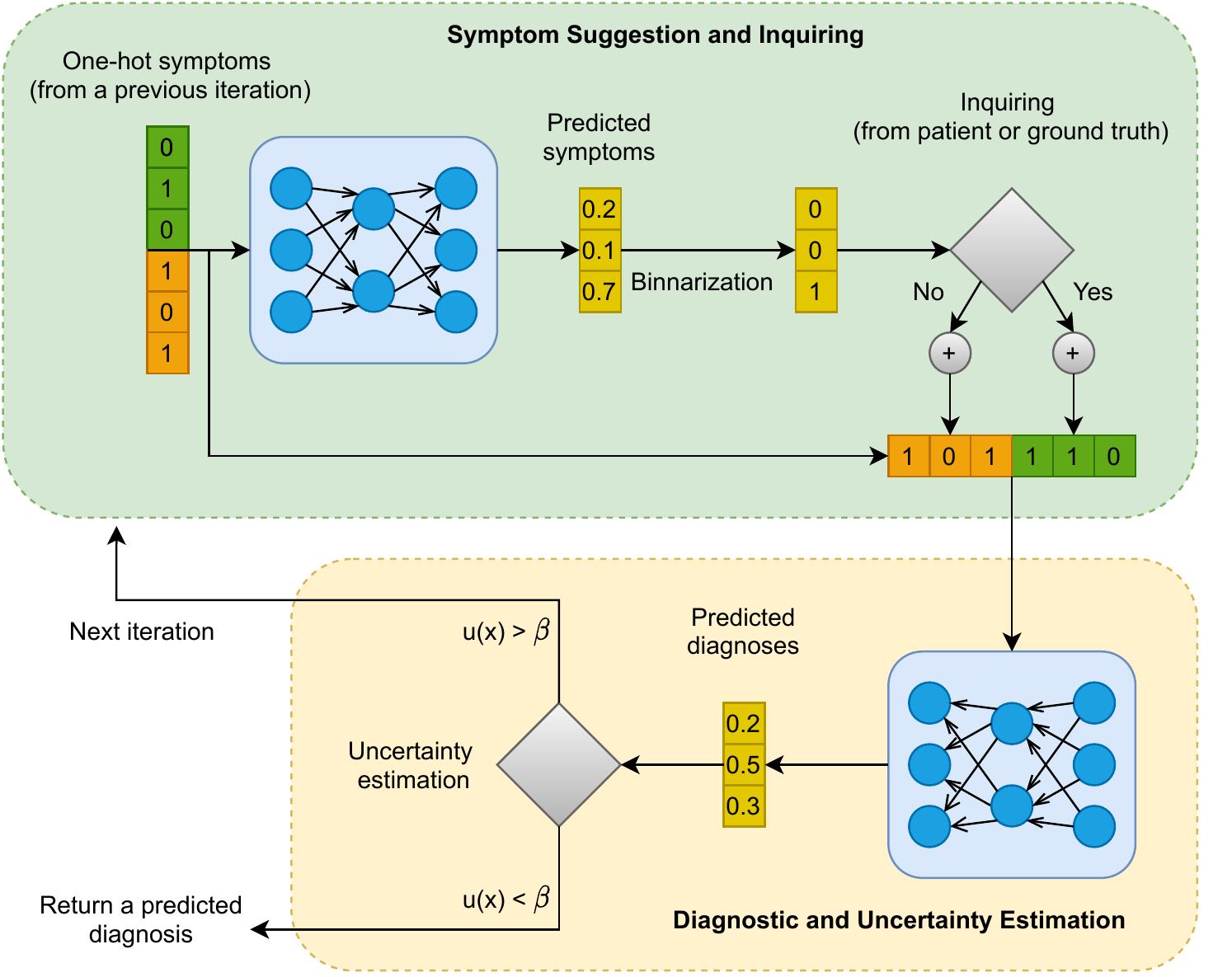}
}

}
\vspace{-0.1cm}
\caption{The architecture of the symptom checker model}
\label{fig:arch}

\vspace{-0.5cm}
\end{figure}

\section{Symptom Checker Model}

We propose a symptom checker model, NeuralSympCheck, consisting of two neural submodels: a network for suggesting symptoms that should be inquired from a patient and a model for performing actual diagnostics. The architecture of the model is presented in Figure \ref{fig:arch}. 

The symptom checker works iteratively. On each iteration, it receives a set of already known symptoms (as well as the information about the absence of some symptoms) and tries to guess the most probable symptom of a patient with the symptom suggestion submodel. The factual information about the presence of the corresponding symptom is inquired from a patient. Then known symptoms and the factual information about the presence/absence of the suggested symptom are used by the diagnostics submodel to predict the disease. We quantify the uncertainty of this prediction, and if it is intolerably high, we start a new iteration of symptom clarification, in which the symptom suggestion submodel receives extended information about symptoms. We note that despite splitting the whole model into two submodels during the inference, they are trained jointly end-to-end with a logic regularization mechanism: the diagnostics submodel learns how to correctly predict diseases with limited information, while the symptom suggestion submodel learns to suggest the most crucial evidence for diagnostics.

\subsection{Symptom Suggestion and Diagnostics Submodels}

The symptom suggestion submodel receives two vectors that encode information known so far about the symptoms. The first vector accumulates the information about present symptoms stored using one-hot encoding, while the second vector stores one-hot encoded information about known absent symptoms. 

The architecture of the submodel is a feed-forward neural network, in which each linear layer is followed by batch normalization, a dropout layer, and a ReLU activation. The model's output is a probability distribution of possibly present symptoms obtained via the softmax function. 

The most probable symptom is queried from the patient (during training, it is taken from the gold standard). Then, the actual information about the presence or absence of the symptom is added to the corresponding %one-hot encoded 
vectors.

The diagnostics submodel takes the extended available information about symptoms as input and predicts a patient's diagnosis. The symptoms are encoded in the same way as the input for the symptom suggestion submodel and the architecture is also the same. The output of the diagnostics model is a probability distribution of potential diseases obtained via the softmax function.

\subsection{Training with Logic Regularization}

During training, we perform the same iterative process of symptom suggestion and diagnosis prediction until the uncertainty of the latter is not low enough. The training of both submodels is performed end-to-end, so the gradient from the diagnostics submodel is propagated into the symptom suggestion submodel. Since the diagnostics submodel takes as input discrete data encoded in one-hot vectors instead of differentiable softmax distributions, the straightforward stacking of these two submodels requires indifferentiable operations. To mitigate this problem and train submodels with the standard backpropagation algorithm, we use a simplified implementation of the Gumbel-softmax approximation \cite{jang2016categorical} without stochastic sampling. 

The overall training loss $L$ is combined from two components: the symptom prediction loss $L_{s}$ and the diagnosis prediction loss $L_{d}$: $L = \lambda L_s + L_d$, where $\lambda > 0$ is a hyperparameter. %The first component 
$L_s$ is the Asymmetric loss \cite{ridnik2021asymmetric} designed for multi-label classification to mitigate the data skewness towards particular classes:
\begin{equation}
L_s=\left\{\begin{array}{l}
L_{+}=(1-p)^{\gamma_{+}} \log (p) \\
L_{-}=\left(p_{m}\right)^{\gamma_{-}} \log \left(1-p_{m}\right),
\end{array}\right.
\end{equation}
where $p$ is a symptom prediction probability, $\gamma_+, \gamma_-$ are focusing hyperparameters, $p_{m} = max(p - m, 0)$, $m \geq 0$ is a margin hyperparameter. $L_d$ is a simple cross-entropy loss commonly used for the standard multi-class classification.

The suggested approach to training these two submodels lies in the paradigm of analytic-synthetic logic regularization. In the analytical approach, a complex model is trained as a sequence of small independent architectures, which improves the interpretability of the solution. In contrast, the synthetic approach trains a single model on a significant target problem (end-to-end), which increases the solution's flexibility. In this paper, none of the approaches described can meet the essential requirements for symptom suggestion. Firstly, symptom suggestion cannot be viewed as task-independent of diagnosis prediction since the goal of this step is not to propose the most likely symptom but the symptom that would potentially reduce the uncertainty of the second submodel the most. Secondly, with the standard end-to-end learning, the information necessary for interpreting and identifying symptoms is lost.

The proposed architecture, in which the predictions of the first submodel are fed to the input of the second submodel and the gradients from the second submodel are propagated to the first submodel, is an attempt to encapsulate both approaches within a single architecture and take advantage of the benefits of each. Such an analytic-synthetic system benefits from two-way regularization: using an explicit symptom prediction subproblem to solve the diagnosis detection problem and using the end goal of the whole problem to regularize the symptom prediction solutions. Because the proposed architecture, in a sense, imitates the logic of decision-making by physicians in real life (additional cyclic tests until the diagnosis is certain), the proposed framework can be considered as one of the variants of logic regularization \cite{riegel2020logical,zhou2020clinical}.

\subsection{Uncertainty Estimation of the Diagnostics Submodel}
\label{sec:uediag}

Following Lin et al. \cite{lin2020learning}, we quantify the uncertainty of the diagnostics submodel and use it as a criterion for stopping ``questioning'' a patient about additional symptoms. This resembles conducting the diagnostics in real life, as a physician collects more evidence only until he is sure enough about the diagnosis. Furthermore, different diseases require a different amount of evidence to make a reliable conclusion, as some diseases are more ambiguous than others. Therefore, exhaustive questioning or asking a fixed considerable number of symptoms is impossible since we would like to make a reliable conclusion as soon as possible, saving time and effort of patients. This also helps speed up training and prevents overfitting, which eventually leads to better performance of the diagnostics submodel. 

In this work, for quantifying uncertainty $u$ of a diagnosis $d$ for a case $x$, we rely on the entropy of the diagnostics submodel output distribution $p(d|x)$ obtained with softmax: $u(x) = E_{p(d|x)}[-\log p(d|x)]$.

We ask for more symptoms until uncertainty of a disease prediction becomes lower than a predefined threshold: $u(x) < \beta, \beta \in (0, 1)$ or we exceed a predefined maximum number of attempts $Q$. The values $Q$ and $\beta$ are hyperparameters that are selected using a validation dataset.

\section{Experiments}

\subsection{Data}

\subsubsection{Real world data.}
The \textit{MuZhi} dataset was created by Zhongyu et al. \cite{wei2018task} from real dialogues on the Chinese healthcare internet portal\footnote{\url{https://muzhi.baidu.com/}}. This dataset encompasses 66 symptoms and four diseases. The dataset consists of 710 records containing the raw dialogue and normalized symptoms checked during the dialogue, either found or not. The symptoms from each record are tagged either as explicit or implicit. The explicit symptoms are the symptoms initially presented by the patient before the beginning of the dialogue. The presence or absence of implicit symptoms is discovered during the recorded dialogue.

The \textit{Dxy} Dialogue Medical \cite{xu2019end} dataset is based on dialogues from a popular Chinese medical forum\footnote{\url{https://dxy.com/}}. It consists of 527 unique dialogues, five diseases, and 41 symptoms. The symptoms from each record are tagged either as explicit or implicit as in the MuZhi dataset.

\vspace{-0.3cm}
\subsubsection{Synthetic data.}
The MuZhi and Dxy datasets are limited in the number of symptoms and diseases. To check the performance of our model in the case of significant symptom and diseases spaces, we used a synthetic dataset \textit{SymCat} presented in \cite{kao2018context} with modifications from \cite{peng2018refuel}. This dataset is created from the similarly named symptom and disease database SymCat \footnote{\url{http://www.symcat.com/}}. It contains information about 474 symptoms and 801 related diseases. 

The dataset is built following the procedure: select a disease from the list; select the symptoms from aposteriori distribution using a Bernoulli experiment for each symptom; split the symptoms into implicit and explicit groups as in the MuZhi and Dxy datasets. As in previous works \cite{peng2018refuel,he2020fit}, to evaluate the system performance on different scales, we use three versions of the dataset with the varying number of diseases -- 200, 300, and 400. We note that we did not find the source code of the generating procedure used in \cite{peng2018refuel,he2020fit}. Therefore, although we reproduced the generation process according to their description, there might be minor deviations. 
Dataset statistics is presented in Table \ref{datasets} in Appendix \ref{appendix:A}.

\subsection{Experimental Setup}

\subsubsection{Hyperparameters and Training Details.}

Model and training hyperparameters, including the number of hidden layers, dropout ratio, layer size, learning rate, number of epochs, scaling coefficient of the multi-label loss, and the uncertainty threshold are selected on the validation datasets using the Optuna package \footnote{\url{https://optuna.org}}. To reduce optimization search space, we use the same number of layers with the same size in both submodels. The selected values are presented in Table \ref{hyperparams} in Appendix \ref{appendix:A}. Training is performed using the corrected version of Adam  with linear decay of the learning rate and warm-up. The focusing hyperparameters are fixed: $\gamma^+=1,\gamma^-=4$. The maximum number of attempts $Q=50$.

\vspace{-0.4cm}
\subsubsection{Evaluation Metrics.}
The quality of disease prediction is evaluated using the top-$k$ accuracy metric $\mathrm{Acc@k}$ ($k \in {1,2,3}$). For each example, if a true disease is present among the top $k$ predictions in the output probability distribution of the model, it is considered as the correct answer of the model. In ablation studies, we also use weighted macro F1 to evaluate symptom prediction quality.

\vspace{-0.4cm}
\subsubsection{Baselines.}
We compare the proposed NeuralSympCheck model with several models from the previous work \cite{peng2018refuel,xu2019end,xia2020generative,zhao2020weighted,he2020fit,guan2021bayesian} and with two simple baselines based on a feedforward neural network. These baselines have the same architecture as submodules: several fully-connected layers with batch normalization, ReLU activation, and dropout regularization. The first baseline performs multi-label disease classification using only the starting explicit symptoms (baseline ex). The second baseline uses both explicit and implicit symptoms (baseline ex\&im), which is unrealistic and very strong assumption.

\subsection{Results and Discussion}

\begin{table}[t]
\centering
\caption{Diagnostics Acc@1 (\%) on the \emph{MuZhi} and \emph{Dxy} datasets} \label{mz_dxy_results}
\vspace{-0.2cm}
\begin{tabular}{r c c} 

\multicolumn{1}{r}{\textbf{Model}} & {\bf MuZhi} & {\bf Dxy} \\
\hline

\multicolumn{1}{r}{Baseline ex} & 61.3 & 66.4 \\
\multicolumn{1}{r}{Baseline ex\&im} & 65.8 & 77.3 \\
\hline

\multicolumn{1}{r}{Peng at al.\cite{peng2018refuel}} & 71.8 & 75.7 \\
\multicolumn{1}{r}{Xu at al.\cite{xu2019end}} & 73.0 & 74.0 \\
\multicolumn{1}{r}{Xia at al.\cite{xia2020generative}} & 73.0 & 76.9 \\
\multicolumn{1}{r}{Zhao at al.\cite{zhao2020weighted}} & 69.7 & 74.0 \\
\multicolumn{1}{r}{He at al.\cite{he2020fit}} & 72.6 & {\bf 81.1} \\
\multicolumn{1}{r}{Guan at al.\cite{guan2021bayesian}} & 65.5 & 80.8 \\
\hline

\multicolumn{1}{r}{Our best results} & {\bf 74.5} & 75.7 \\
\hline
\end{tabular}
\vspace{-0.05cm}
\end{table}

Table \ref{mz_dxy_results} presents the main experimental results on small datasets MuZhi and Dxy. On the MyZhi dataset, our NeuralSympCheck model achieves the new state-of-the-art, outperforming all the baselines and the models from the previous work. On the Dxy dataset, our model outperforms the first baseline and most of the systems from the previous work, only falling behind the recently proposed RL-based systems presented in \cite{xu2019end,he2020fit,guan2021bayesian}. We attribute this to the fact that Dxy is smaller, contains less number of symptoms, and has a smaller average number of implicit symptoms that can be clarified for the final diagnosis.

\begin{table}[t]
\centering
\caption{Results (\%) on the test part of the \emph{SymCat} datasets} \label{symcat_results}
\vspace{-0.2cm}
\scalebox{0.92}{
\begin{tabular}{r | c c c | c c c | c c c} 

\multicolumn{1}{c} {} & \multicolumn{3}{c} {\textbf{200 diseases}} & \multicolumn{3}{c} {\textbf{300 diseases}} & \multicolumn{3}{c} {\textbf{400 diseases}} \\

{\textbf{Model}} & \textbf{Acc@1} & \textbf{Acc@3} & \textbf{Acc@5} & \textbf{Acc@1} & \textbf{Acc@3} & \textbf{Acc@5} & \textbf{Acc@1} & \textbf{Acc@3} & \textbf{Acc@5} \\
\hline

{Baseline ex} & 46.7 & 70.6 & 81.1 & 41.1 & 63.0 & 73.4 & 36.3 & 57.0 & 67.8 \\
{Baseline ex\&im} & 82.6 & 96.7 & 99.1 & 78.4 & 94.0 & 98.0 & 74.4 & 92.0 & 97.0 \\
\hline

{Peng at al.\cite{peng2018refuel}} & 54.8 & 73.6 & 79.5 & 47.5 & 65.1 & 71.8 & 43.8 & 60.8 & 68.9 \\
{He at al.\cite{he2020fit}} & 55.6 & 80.7 & 89.3 & 48.2 & 73.8 & 84.2 & 44.6 & 69.2 & 79.5\\
\hline

{Our best results} & {\bf 63.2} & {\bf 89.3} & {\bf 96.7} & {\bf 54.8} & {\bf 81.2} & {\bf 91.1} & {\bf 49.8} & {\bf 76.6} & {\bf 87.9} \\
\hline
\end{tabular}
}
\vspace{-0.3cm}

\end{table}

As we can see from Table \ref{symcat_results}, on the more extensive synthetic datasets based on SymCat, NeuralSympCheck also achieves state-of-the-art results, outperforming all previous models and the first baseline. The best results are achieved for each number of possible diagnoses. Our model does not reach the performance of the second unrealistic baseline trained on both explicit and implicit symptoms. This may happen because NeuralSympCheck overcomes the uncertainty threshold early and stops clarifying additional symptoms. We note that our solution outperforms models from the previous work with a significant margin. We attribute this remarkable achievement to using a conceptually novel model architecture that is better adapted to the big and sparse symptom space.

\subsection{Ablation Studies}

The goal of the first ablation study is to evaluate the effect of the symptom prediction loss (Table \ref{tab:ablation} in Appendix \ref{appendix:B}). We train our model using only the first diagnosis prediction loss with a fixed number of clarification iterations. This helps to improve Acc@1 of the diagnostics model. However, this results in a substantial reduction of symptom-suggestion model F1 compared to training with both losses. We conclude that training only with the diagnosis classification loss facilitates the symptom suggestion submodel to exceedingly adjust its predictions in the direction of best coherence with the predicted diagnosis.

In the second ablation study, we evaluate the effect of uncertainty estimation. Table \ref{tab:ablation} in Appendix \ref{appendix:B} shows that models using uncertainty estimation achieve the best results in terms of the Acc@1 metric for diagnosis prediction. However, the F1 metric for symptom prediction is significantly lower. The obtained results can be explained by the fact that models using uncertainty estimation conduct fewer symptom clarification iterations that are only necessary to achieve model confidence in the correct diagnosis.

We also test the hypothesis that additional symptoms help to reduce the uncertainty of the diagnosis submodel predictions. Figure \ref{fig:entropy_per_iter} in Appendix \ref{appendix:B} shows that, indeed, regardless of the dataset, the more iterations of symptom refinement are performed, the less uncertain the diagnostics submodel predictions are.

\section{Conclusion}

We presented a novel model for symptom and diagnosis prediction based on supervised learning. It outperforms recently proposed RL-based counterparts and mitigates some of their limitations, such as the complexity of learning, the fundamental flaws of the Markov process, and the complexity of applying RL-based methods in practice. By leveraging asymmetric loss, we overcome the problem of large and sparse symptoms space. We propose an approach that allows training the symptom suggestion and the diagnosis prediction models in an end-to-end fashion with standard backpropagation. We are the first to use logic regularization for the considered task, which effectively helps to predict relevant symptoms for diagnostics. Finally, uncertainty estimation of diagnosis prediction is used as a stopping criterion for asking about new symptoms. Our NeuralSympCheck model achieves the new state of the art on datasets with large symptom and diagnosis spaces.

We want to emphasize the practical significance of this work because the presented model is relatively easy to implement, stable in training, and not demanding on computational resources. This makes it possible to apply the proposed model in real-world medical systems, which is our future work direction. 

\vspace{-0.5cm}

\section*{Acknowledgements}
We are grateful to anonymous reviewers for their valuable feedback. The work was supported by the RSF grant 20-71-10135. 

\bibliography{bibliography.bib}

\clearpage
\appendix

\section{Dataset Statistics and Hyperparameters} 
\label{appendix:A}

\vspace{-0,8cm}

\begin{table}[]
\footnotesize
\centering
\caption{Dataset statistics} \label{datasets}
\vspace{-0.2cm}
\scalebox{0.83}
{
\begin{tabular}{r c c c c c} 

\multicolumn{1}{r}{} & {\bf MuZhi} & {\bf Dxy} & {\bf SymCat 200} & {\bf SymCat 300} & {\bf SymCat 400} \\
\hline

\multicolumn{1}{r}{Total dialogues} & 710 & 527 & 1,110,000 & 1,110,000 & 1,110,000 \\
\multicolumn{1}{r}{Training dialogues} & 568 & 423 & 1,000,000 & 1,000,000 & 1,000,000 \\
\multicolumn{1}{r}{Validation dialogues} & - & - & 100,000 & 100,000 & 100,000 \\
\multicolumn{1}{r}{Testing dialogues} & 142 & 123 & 10,000 & 10,000 & 10,000 \\
\hline
\multicolumn{1}{r}{Unique diagnoses} & 4 & 5 & 200 & 300 & 400 \\
\multicolumn{1}{r}{Unique symptoms} & 66 & 41 & 326 & 350 & 367 \\
\hline
\multicolumn{1}{r}{Average number of explicit symptoms} & 2.4 & 3.1 & 1.9 & 2.0 & 2.0 \\
\multicolumn{1}{r}{Average number of implicit symptoms} & 2.4 & 1.2 & 1.9 & 2.0 & 2.0 \\
\hline
\end{tabular}
}
\end{table}

\vspace{-1.0cm}

\begin{table}[]
\footnotesize
\centering
\caption{Hyperparameters of the models that showed the best results on validation datasets} \label{hyperparams}
\vspace{-0.3cm}
\scalebox{0.95}{
\begin{tabular}{r c c c c c} 

\multicolumn{1}{r}{\textbf{Hyperparams}} & {\bf MuZhi} & {\bf Dxy} & {\bf SymCat 200} & {\bf SymCat 300} & {\bf SymCat 400}\\
\hline

\multicolumn{1}{r}{Size of the first layer} & 6,000 & 10,000 & 8,000 & 8,000 & 8,000 \\
\multicolumn{1}{r}{Size of the second layer} & 3000 & - & - & - & - \\
\multicolumn{1}{r}{Dropout probability} & 0.4 & 0.5 & 0.5 & 0.5 & 0.5 \\
\multicolumn{1}{r}{Multilabel loss coefficient} & 1.6 & 0.6 & 1 & 1 & 1 \\
\multicolumn{1}{r}{Minimum uncertainty value, $\beta$} & 0.5 & 0.3 & 0.3 & 0.4 & 0.4 \\
\multicolumn{1}{r}{Number of epochs} & 35 & 5 & 5 & 10 & 10 \\
\multicolumn{1}{r}{Learning rate} & 5e-5 & 1e-3 & 1e-3 & 1e-2 & 1e-2 \\
\hline

\end{tabular}
}

\end{table}

\vspace{-0.8cm}

\section{Additional Experimental Results}
\label{appendix:B}

\vspace{-1.cm}

\begin{figure}[ht!]
\center{\includegraphics[scale=0.21]{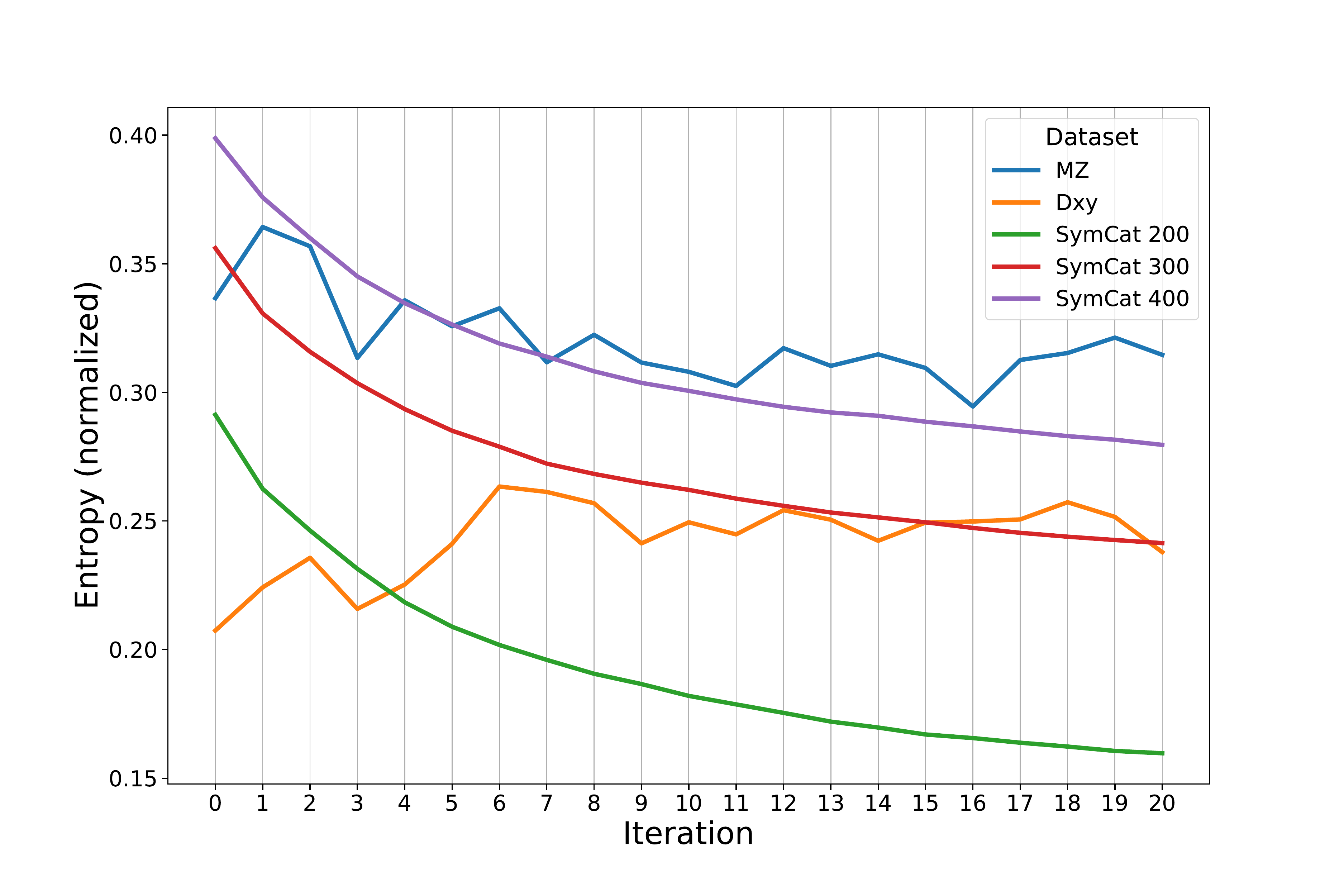}}
\vspace{-0.5cm}
\caption{Change of entropy value depending on iteration of symptom inquiring}
\label{fig:entropy_per_iter}
\vspace{-1.1cm}
\end{figure}

\begin{table}[ht!]
\centering
%\scriptsize
\footnotesize
\caption{Ablation studies results (\% Acc@1 by diagnosis / F1 weighted by symptoms)} \label{tab:ablation}
\vspace{-0.2cm}
\scalebox{0.9}{
\begin{tabular}{r| c | c| c| c| c} 

\multicolumn{1}{r}{} & {\bf MuZhi} & {\bf Dxy} & {\bf SymCat 200} & {\bf SymCat 300} & {\bf SymCat 400} \\
\hline

\multicolumn{1}{r}{Only diagnosis loss} & 67.2 / 32.7 & 71.9 / 24.0 & 70.7 / 23.7 & 64.1 / 24.5 & 57.6 / 24.4 \\
\multicolumn{1}{r}{Two losses, without entropy} & 70.3 / 10.2 & 69.0 / 32.8 & 59.6 / 35.0 & 53.8 / 33.2 & 47.7 / 32.3 \\
\multicolumn{1}{r}{With entropy} & 68.3 / 28.4 & 69.1 / 19.2 & 63.2 / 20.7 & 54.8 / 15.8 & 49.8 / 14.5 \\
\hline
\end{tabular}
}

\end{table}

\vspace{-0.5cm}

\end{document}